\documentclass{article}

\usepackage[preprint]{neurips_2025}

\usepackage[utf8]{inputenc} 
\usepackage[T1]{fontenc}    
\usepackage{hyperref}       
\usepackage{url}            
\usepackage{booktabs}       
\usepackage{amsfonts}       
\usepackage{nicefrac}       
\usepackage{microtype}      
\usepackage{xcolor}         

\usepackage[utf8]{inputenc} 
\usepackage[T1]{fontenc}    
\usepackage{hyperref}       
\usepackage{url}            
\usepackage{booktabs}       
\usepackage{amsfonts}       
\usepackage{nicefrac}       
\usepackage{microtype}      

\usepackage{amsmath} 
\usepackage{mathtools} 
\usepackage{amssymb} 
\usepackage{tabularx} 
\usepackage{multirow}

\usepackage{cleveref}       
\usepackage{lipsum}         
\usepackage{graphicx}
\usepackage{natbib}
\usepackage{doi}

\usepackage{natbib} 
\usepackage{mathtools} 
\usepackage{booktabs} 
\usepackage{tikz} 

\usepackage[ruled,vlined]{algorithm2e}

\usepackage{array}

\usepackage{subcaption}
\usepackage{wrapfig}

\title{Bayesian Inverse Problems Meet Flow Matching: Efficient and Flexible Inference via Transformers}

\author{%
  Daniil~Sherki \\
    Skolkovo Institute of Science and Technology\\
    Sberbank, AI4S Center \\
    Moscow, Russian Federation \\
    \texttt{daniil.sherki@skoltech.ru}
    \And
    Ivan~Oseledets \\
    Skolkovo Institute of Science and Technology\\
    Artificial Intelligence Research Institute \\
    Moscow, Russian Federation \\
    \texttt{i.oseledets@skoltech.ru} \\
    \And
    Ekaterina~Muravleva \\
    Skolkovo Institute of Science and Technology\\
    Sberbank, AI4S Center \\
    Moscow, Russian Federation \\
    \texttt{e.muravleva@skoltech.ru} \\
}

\begin{document}

\maketitle

\begin{abstract}
The efficient resolution of Bayesian inverse problems remains challenging due to the high computational cost of traditional sampling methods. In this paper, we propose a novel framework that integrates Conditional Flow Matching (CFM) with a transformer-based architecture to enable fast and flexible sampling from complex posterior distributions. The proposed methodology involves the direct learning of conditional probability trajectories from the data, leveraging CFM’s ability to bypass iterative simulation and transformers’ capacity to process arbitrary numbers of observations. The efficacy of the proposed framework is demonstrated through its application to three problems: a simple nonlinear model, a disease dynamics framework, and a two-dimensional Darcy flow Partial Differential Equation. The primary outcomes demonstrate that the relative errors in parameters recovery are as low as 1.5\%, and that the inference time is reduced by up to 2000 times on CPU in comparison with the Monte Carlo Markov Chain. This framework facilitates the expeditious resolution of Bayesian problems through the utilisation of sampling from the learned conditional distribution.
\end{abstract}


\section{Introduction}
\label{sec:intro}
Many natural processes can be mathematically modeled using appropriate formal representations. However, the challenge often lies in inferring latent parameters that are not directly observable. These parameters must typically be estimated from limited observations, giving rise to Bayesian inverse problems.  The idea of Bayesian inversion is to parametrize the posterior distribution of model parameters, given observations and a prior distribution on the model parameters. The main challenge is that typically the distribution is known up to a normalization constant, making sampling from the posterior intractable. Classical methods like Markov Chain Monte Carlo (MCMC) \citep{geyer1992practical} rely on many forward problem solutions for each set of observations, which can be very time-consuming.

The Bayesian inversion is widely used for addressing inverse problems across diverse domains such as physics and engineering \citep{Cotter2009Bayesian, koval2024tractableoptimalexperimentaldesign}. Its appeal lies in its ability not only to deliver a solution estimate but also to quantify the associated uncertainty. Understanding the distribution of a computed quantity is particularly valuable in applications like digital twins \citep{Kapteyn2021-gp}. For instance, one may need to recover parameters of an ODE system modeling disease spread from observed infection data, or reconstruct permeability fields from indirect measurements \citep{koval2024tractableoptimalexperimentaldesign}. 

A natural approach for tackling Bayesian inverse problems is to apply generative models. There are many available options, like variational autoencoders \citep{kingma2022autoencodingvariationalbayes}, Generative Adversarial Networks (GAN) \citep{goodfellow2014generativeadversarialnetworks} or diffusion models \citep{sohldickstein2015deepunsupervisedlearningusing}, normalizing flows offer exact likelihood estimation \citep{pmlr-v244-gudovskiy24a} while avoiding these computational bottlenecks.
In this work, we focus on a recent generative modelling technique, conditional flow matching, and show that it can be efficiently and easily applied to different Bayesian inverse problems.

\paragraph{Our contribution} 
\begin{itemize}
    \item We formulate the Bayesian inverse problem as the problem of learning the conditional probability distribution from samples, that can be easily constructed.
    \item We propose a transformer-based Conditional Flow Matching (CFM) \cite{lipman2023flowmatchinggenerativemodeling} architecture that can handle different number of observations.
    \item We test our method on several inverse problems and compare it to the MCMC approach.
\end{itemize}

\section{Background and Related Work}
\label{sec:background}
\paragraph{Classical Approaches to Solving Bayesian Inverse Problems}
                                                            
The primary challenges associated with classical methods for solving Bayesian Inverse Problems include high computational costs, difficulties with variational methods, the need for numerous evaluations of the forward model, and limitations in real-time inversion capabilities. Specifically, sampling from complex Bayesian posterior distributions using statistical simulation techniques, such as Markov Chain Monte Carlo (MCMC), Hamiltonian Monte Carlo, and Sequential Monte Carlo, is computationally expensive. Variational inference algorithms, including mean-field variational inference and Stein variational gradient descent, face challenges in high-dimensional settings due to the difficulty of accurately approximating posterior distributions. Additionally, these methods require multiple evaluations of the forward model and complicated parametric derivatives, further increasing computational costs in high-dimensional scenarios. Consequently, classical approaches may be less efficient for real-time inversions, particularly when dealing with new measurement data, highlighting the need for more efficient alternatives such as deep learning-based methods \cite{guan2023efficientbayesianinferenceusing}.

\paragraph{Deep Learning Models for Solving Inverse Problems}

Bayesian inverse problems have also been addressed using physics-informed neural networks \citep{RAISSI2019686} by combining invertible flow-based neural networks (INNs). In \citep{guan2023efficientbayesianinferenceusing}, this approach is shown to be effective but not universally applicable, as it requires designing a custom loss function for each PDE to ensure efficient training. Additionally, the number of observations in the Darcy flow problem is fixed.

A significant number of inverse problems in medicine can be effectively addressed using generative networks \citep{Aali_2023, song2022solvinginverseproblemsmedical}. In \cite{simformer2024allinone} proposed Simformer, a framework for simulation-based inference that combines transformer architectures with score-based diffusion models.  Simformer relies on stochastic diffusion processes and does not directly learn deterministic transport maps, which can limit interpretability and inference speed in structured inverse problems. 

\paragraph{Generative Adversarial Networks and Variational Autoencoders}
Generative Adversarial Networks (GANs), first introduced by \cite{goodfellow2014generativeadversarialnetworks}, have become a cornerstone of generative modeling. Recent advances demonstrate their applicability to Bayesian inverse problems. For instance, \cite{mücke2022markovchaingenerativeadversarial} proposed MCGAN, a GAN-based framework to circumvent the computational burden of traditional MCMC methods. By replacing the physical forward model with a trained generator during inference, their approach accelerates likelihood evaluations for complex PDE-based problems.

GANs have also been integrated into hybrid Bayesian frameworks. In \cite{Patel2020bayesianinferenceinphysics}, a GAN was employed to approximate high-dimensional parameter priors within MCMC sampling. Similarly, \cite{Xia_2022} combined a VAE-based prior with MCMC for posterior estimation. This method was called Multiscale deep generative model (MDGM). Although these methods leverage generative models to enhance prior representation, their computational gains remain limited, as they still require iterative forward model evaluations during sampling. Other works, such as \cite{goh2021solvingbayesianinverseproblems} using VAEs embed generative models into variational inference frameworks. However, these approaches face trade-offs between approximation accuracy and scalability in high-dimensional settings.

However these approaches have two fundamental disadvantages. First, they cannot handle an arbitrary number of observations, which can be critical in real-world problems. Second, these approaches often do not escape the iterative process itself due to the fact that such approaches do not explicitly generate a posterior distribution.

\paragraph{Flow Matching}

Flow Matching (FM) \citep{lipman2023flowmatchinggenerativemodeling} is an efficitient approach to generative modeling based on Continuous Normalizing Flows (CNFs) \citep{NEURIPS2018_ChenCNFs}, enabling large-scale CNF training without simulation by regressing vector fields of fixed conditional probability paths. It generalizes diffusion models by supporting a broader class of Gaussian probability paths, including Optimal Transport (OT) displacement interpolation, which enhances efficiency, stability, and sample quality. Compared to diffusion-based methods, FM allows faster training and sampling, improves likelihood estimation, and enables reliable sample generation using standard numerical ODE solvers, making it an alternative for high-performance generative modeling.

\cite{whang2021composingnormalizingflowsinverse} demonstrated their utility in Bayesian inverse problems by embedding a Normalizing Flow into a variational inference framework, enabling flexible posterior approximations. While these methods offer theoretical guarantees on invertibility, their computational cost grows with model complexity, limiting their practicality for large-scale physical systems.

\begin{table*}[h]
\centering
\begin{tabular}{@{}llccc@{}}
\toprule
\multicolumn{1}{c}{\textbf{Method}} & \multicolumn{1}{c}{\textbf{Base model} } & \begin{tabular}[c]{@{}c@{}}\textbf{Exact}\\ \textbf{likelihood}\\ \textbf{estimation}\end{tabular} & \begin{tabular}[c]{@{}c@{}}\textbf{No middle-man}\\ \textbf{Training}\end{tabular} & \begin{tabular}[c]{@{}c@{}}\textbf{Arbitrary} \\ \textbf{number of}\\ \textbf{observations}\end{tabular} \\ \midrule
MDGM & VAE based on CNN  & \texttimes & \checkmark  & \texttimes  \\
MCGAN  & MCMC + GAN & \texttimes  & \texttimes & \texttimes  \\
PI-INN & PI + flow-based model & \checkmark   & \checkmark & \texttimes  \\
CFM-Tr (ours) & CFM + Transformer  & \checkmark  & \checkmark &\checkmark \\ \bottomrule
\end{tabular}
\caption{Comparison of methods for solving Bayesian Inverse problems. *MDGM use the PDE solution as a holistic observation; the problem was not formulated as the recovery of the forward model from a small number of observations}
\label{tab:comparison}
\end{table*}

Table \ref{tab:comparison} compares methods for solving Bayesian inverse problems. Deep learning methods for solving Bayesian inverse problems exhibit distinct strengths and limitations. MDGM \citep{Xia_2022} leverages a VAE-based convolution neural network with MCMC for multiscale inference, excelling in high-dimensional PDE-based problems but lacking exact likelihood estimation and flexibility for arbitrary observations. MCGAN \citep{mücke2022markovchaingenerativeadversarial} combines MCMC with GANs for high-fidelity posterior sampling but suffers from computational complexity, lack of explainability, and fixed observation models. PI-INN \citep{guan2023efficientbayesianinferenceusing} employs physics-informed flow-based models, enabling exact likelihood estimation and end-to-end training but struggles with variable observation sizes due to architectural constraints. In contrast, CFM-Tr integrates conditional flow matching with transformers, offering exact likelihood estimation, end-to-end training, and adaptability to arbitrary observations, making it suitable for dynamic inverse problems like real-time medical imaging. While MDGM and PI-INN are effective for structured problems, CFM-Tr addresses key limitations by combining flexibility, exact inference, and scalability.


\section{Methodology}
Consider a forward model defined as: 

\begin{equation*}\label{eq:forward_model}
    d = \mathcal{F}(m, e) + \eta,
\end{equation*}

where $m$ represents model parameters sampled from their prior distribution, $e$ denotes experimental conditions or design parameters, and $\eta$ is random noise sampled from a predefined noise distribution.

The Bayesian inverse problem aims to infer unknown/unobservable parameters \(m\) using known experiment parameters \(e\) and observations \(d\) from the forward model. The solution is characterized by a posterior probability distribution, with density given by Bayes' law:

\begin{equation*}
\pi(m \vert d, e) = \frac{\pi(d \vert m, e)\pi(m)}{\pi(d \vert e)},
\end{equation*}

where $\pi(m)$ is the prior distribution encoding prior knowledge about parameters, $\pi(d \vert m, e)$ is the likelihood, and $\pi(m \vert d, e)$ is the posterior distribution.

The primary objective is to solve the inverse problem: given observations $d$ and experiment parameters $e$, infer the model parameters $m$. Since $m$ is not uniquely determined by $d$ and $e$, it is characterized by the conditional distribution $\pi(m \vert d, e)$. The solution can be reformulated as learning the \textbf{conditional distribution} $\pi(m \vert d, e)$.

To achieve this, we employ the conditional flow matching (CFM) framework from \citep{lipman2023flowmatchinggenerativemodeling} (Algorithm \ref{alg:cfm}). This involves first sampling from an unconditional prior distribution for $m$ (denoted as $m_0$). 

\begin{figure}[ht]
  \centering
  \includegraphics[width=.75\columnwidth]{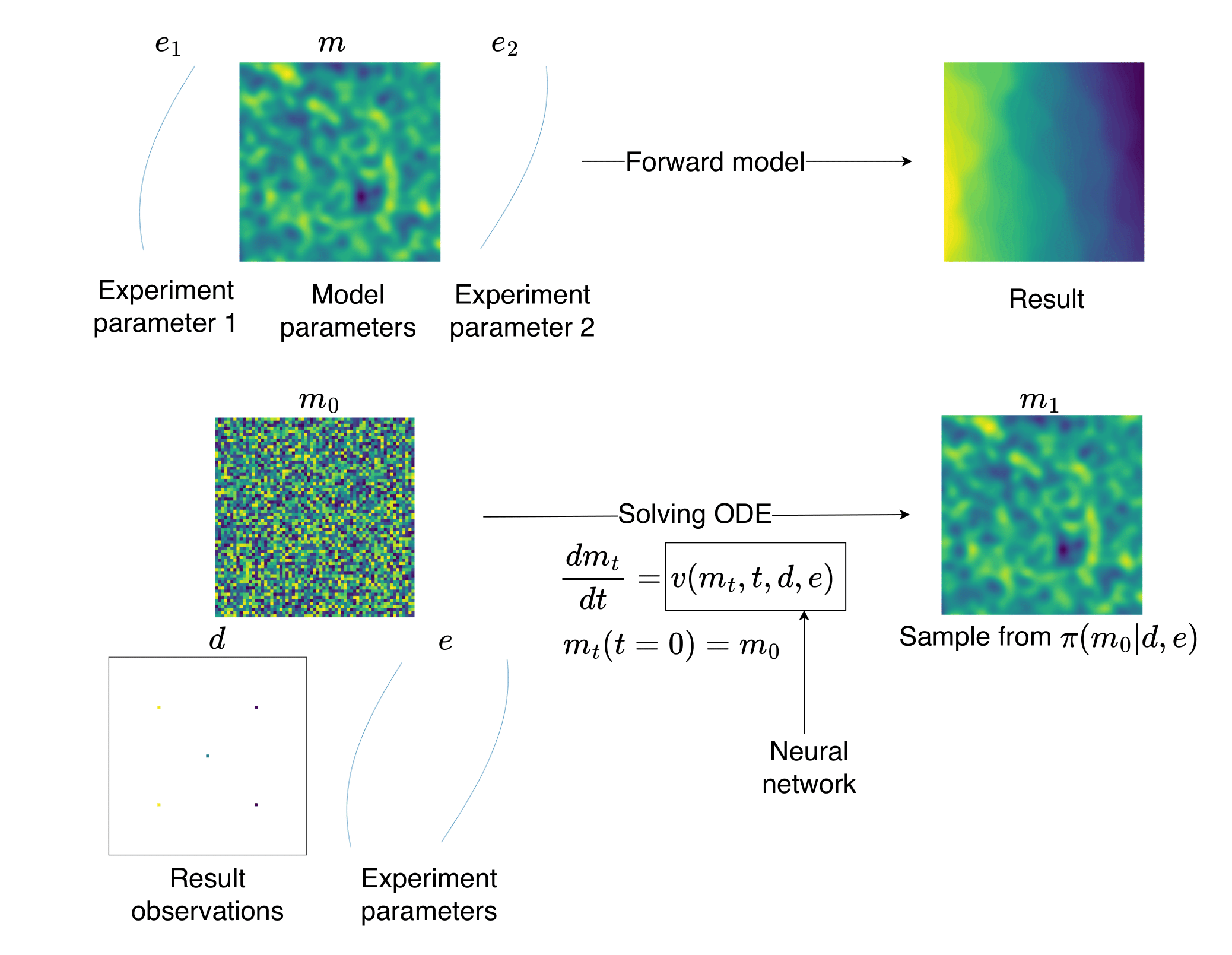}
\caption{Solving the inverse problem using flow-matching scheme}
\label{fig:general_sheme}
\end{figure}

\paragraph{Dataset} The key idea is that we can easily sample from the joint distribution $(m_i, d_i, e_i)$. In order to do that. we generate random model parameters (from the prior distribution) and random observation points. When a pair $m_i, e_i$ is given, we can compute $d_i$ using the forward model. Importantly, $m_i$ is also a sample from the conditional distribution $\pi(m \vert d_i, e_i)$. Thus, when a forward model and prior distributions of model parameters \(m\) and experimental parameters \(e\) are known, we generate training data by sampling multiple variants of \(m\) and \(e\) and computing the forward model to obtain observations \(d\).
For each model parameter $m_i$ we sample $d_i$ for several points $e_i$, thus, our training data consists of tuples of the form of the form $(m_i, d_i, e_i)$, where $d_i$ and $e_i$ may have variable lengths. The model should be able to sample $m_i$ given observations $(d_i, e_i)$. In order to do that, we utilize CFM.

\paragraph{Training} We define a conditional interpolation path between $(m_0, d, e)$ and $(m, d, e)$, where $(d, m, e)$ is drawn from the dataset.



In the CFM approach, we learn a velocity field $v_\theta(m_t, t, d, e)$ that minimizes:


\begin{equation*}
    \mathbb{E}_{t \sim \mathcal{U}(0,1)} 
    \mathbb{E}_{m_0 \sim \text{prior}} 
    \mathbb{E}_{(m, d, e) \sim data} 
    \left[ 
        \big\Vert v_\theta(m_t, t, d, e) - (m - m_0) \big\Vert^2 
    \right] 
    \to \min_\theta
\end{equation*}

where the interpolation path is given by:

\begin{equation*}
    m_t = (1 - t)m_0 + t \cdot m, \quad t \in [0, 1]
\end{equation*}


Here, \(v_\theta\) is a learnable function parameterized by \(\theta\) that predicts the velocity field given inputs \((m_t, t, d, e)\). The input dimensions correspond to \(m_t\), \(t\), \(d\), and \(e\), while the output dimension matches that of \(m\).

During training, elements $(d, m, e)$ are sampled from the dataset, and $m_0$ is drawn from the prior for each iteration of a stochastic optimizer. A neural network effectively represents $v_\theta$ in our experiments.

Once trained, samples from $\pi(m \vert d, e)$ are generated by solving an ordinary differential equation (ODE) parameterized by the learned velocity field.

\begin{wrapfigure}{r}{0.35\textwidth}
  \centering
  \vspace{-10pt}
  \includegraphics[width=\linewidth]{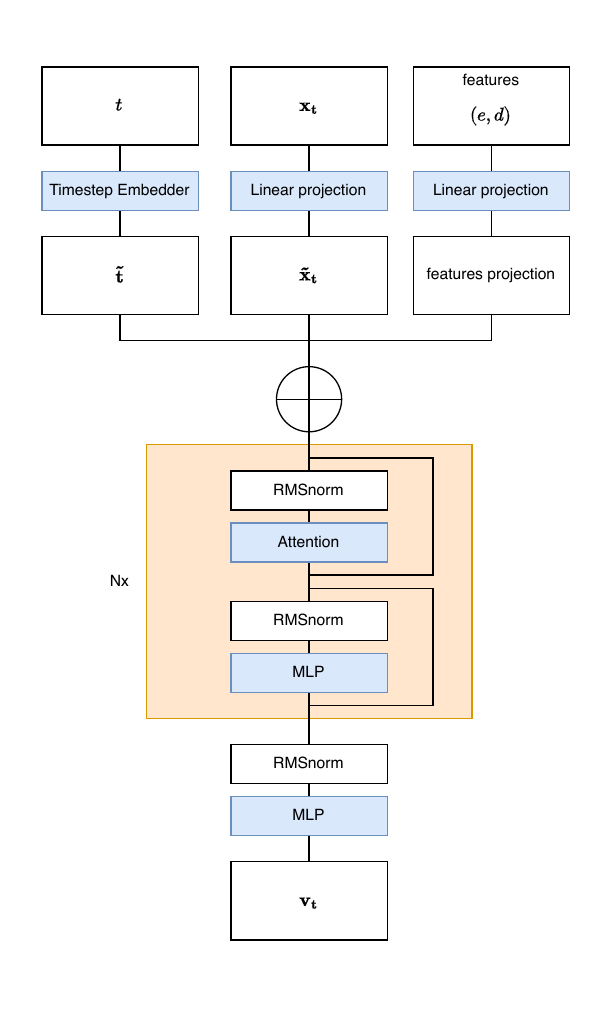}
  \caption{Transformer architecture}
  \label{fig:arch_scheme}
  \vspace{-10pt}
\end{wrapfigure}

A key feature of our approach is the ability to handle arbitrary numbers of observations \(d\) and design parameters \(e\) as input. This capability stems from our transformer architecture, shown in Figure~\ref{fig:arch_scheme}.

\paragraph{Architecture} We parameterize the velocity field using a transformer architecture with bi-directional attention, motivated by the Diffusion Transformer \cite{Peebles2022DiT}. Specifically, our transformer implementation uses linear projection of input parameters into the embedding space. Time is encoded using a Timestep Embedder as proposed in \citep{Peebles2022DiT}, which ensures proper time representation in the embedding space. Root Mean Square (RMS) normalization stabilizes learning dynamics. The activation function is \(x = \text{ReLU}(x)^2\). Self-attention uses rotary position embeddings (RoPE), enabling the transformer to learn relative token positions and generalize to sequences longer than those seen during training.

The architecture varies slightly across tasks to accommodate different input data representations. Specific implementations for tasks from Section \ref{sec:exps} are detailed in Figure \ref{fig:arch_scheme_2} in Appendix \ref{app:tech}.

Model inference follows Algorithm \ref{alg:cfm_inference}, where the trained CFM model serves as the velocity field in the ODE.


\paragraph{Handling variable number of observations} We need to be able to predict the model parameters from different number of observation points. As mentioned before, we generate datasets with varying numbers of observation points, where each batch corresponds to samples with a specific number of points. During training, the model processes batches with different numbers of points sequentially. We propose two strategies: the first follows Algorithm \ref{alg:cfm}, computing gradients for each batch and updating after a fixed number of batches.
\label{sec:method}

\section{Numerical Experiments}
\label{sec:exps}
We utilize numerical experiment formulations adopted from \citep{koval2024tractableoptimalexperimentaldesign}. Specifically, we consider solutions of ordinary differential equation systems modeling disease propagation, as well as elliptic partial differential equations such as the Darcy Flow. These problem classes are widely employed in the literature on Bayesian inverse problems.

\subsection{Simple nonlinear model}

After 10,000 runs of the trained model, the generation error is $1.5 \cdot 10^{-3} \pm 0.9 \cdot 10^{-3}$. Figure \ref{fig:flow_matching_pathes} shows example paths as we move from the prior distribution to the target distribution $\pi(d, m, e)$. Notably, due to the efficient learning of Flow Matching, the paths are almost straight, indicating optimal transport.

\subsection{SEIR disease model}
\label{subsec:seir}

The SEIR (Susceptible-Exposed-Infected-Removed) model is a mathematical framework used to simulate the spread of infectious diseases. In this case study, we simulate a realistic scenario where we measure the number of infected and deceased individuals at random times and use this information to recover the control parameters of the ODE system.

For $\mathbf{m}_{\textbf{true}} = [0.4, 0.3, 0.3, 0.1, 0.15, 0.6]$, after 1,000 calculations the average error is $2.05\% \pm 1.04\%$ using a 4-point multilayer perceptron (MLP) model.

\subsection{Permeability field inversion}
\label{subsec:inv_pde_3point}

We next consider the problem of solving a two-dimensional elliptic PDE. This type of problem is common in the oil industry, where pressure observations from a small number of wells are used to reconstruct the permeability field of an oil reservoir. The equation also has applications in groundwater modeling and many other domains.

Our results show that we can effectively recover the PDE coefficient using just a few strategically placed measurement points. Figure \ref{fig:inv_perm} demonstrates that with 8 relatively uniformly distributed points over the solution field, we can obtain an almost identical solution (approximately 2.75\% relative error). The ensemble-generated $\log \kappa$ represents the mean of 50 parameter predictions from the transformer's inference.

\section{Results and Discussions}
\label{sec:res_disc}
Table \ref{tab:seir_permeability_results} presents the results of numerical experiments for our proposed method using the following error metric:

\begin{equation*}
    \varepsilon = \frac{\Vert \text{DE}(m) - \text{DE}(\tilde{m}) \Vert}{\Vert \text{DE}(m)\Vert}
\end{equation*}

where \(\text{DE}\) represents the solution of the differential equation (ODE or PDE) using either the true parameters \(m\) or the generated parameters \(\tilde{m}\), computed as an average over 10 generations from the flow matching model.

\begin{figure}[!htb]
\centering
\begin{minipage}[t]{0.48\textwidth}
    \centering
    \includegraphics[width=\linewidth]{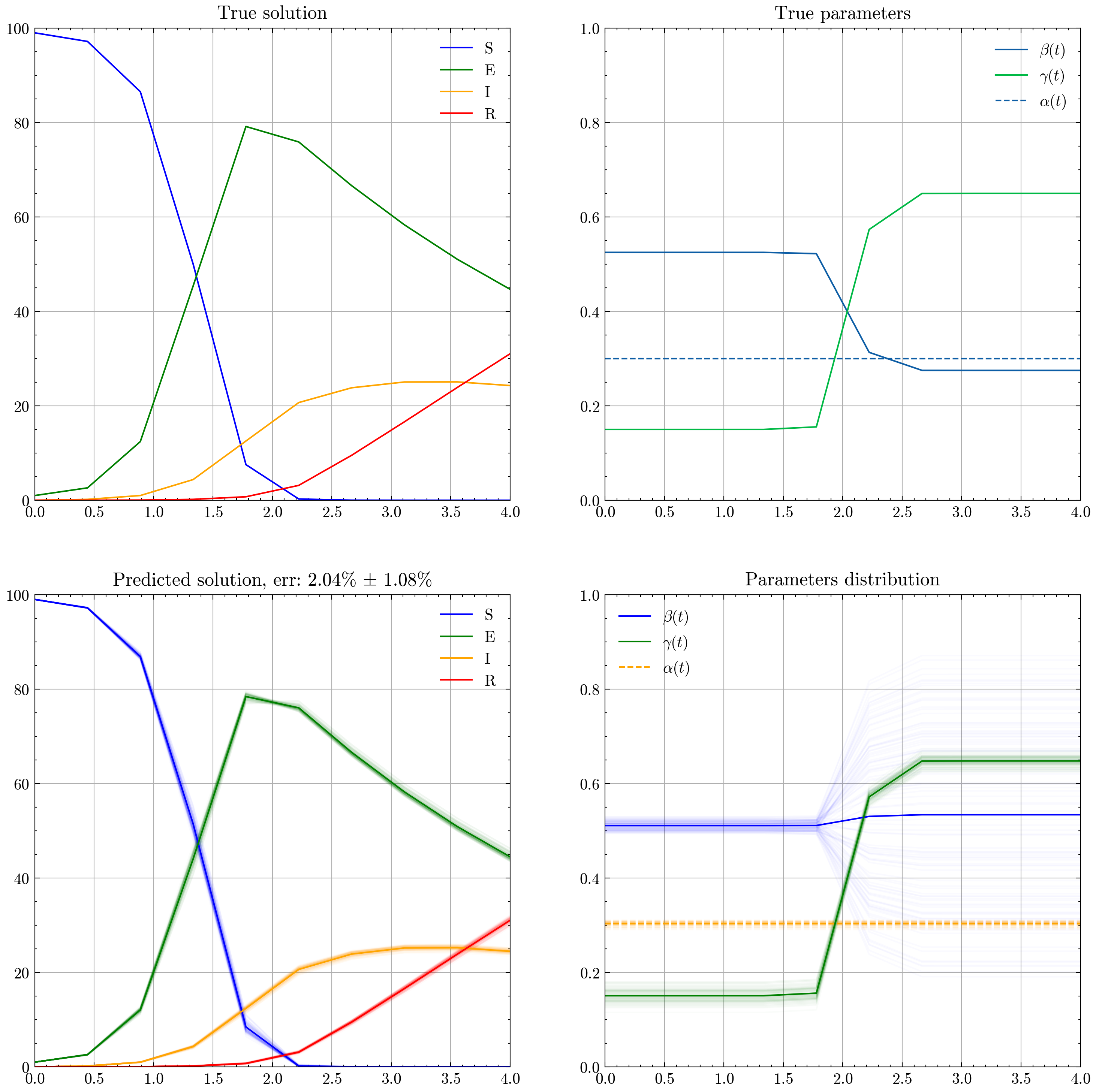}
    \caption{Probabilistic solutions to the inverse problem for $\mathbf{m}_{\textbf{true}} = [0.4, 0.3, 0.3, 0.1, 0.15, 0.6]$}
    \label{fig:odes_sols}
\end{minipage}%
\hfill
\begin{minipage}[t]{0.48\textwidth}
    \centering
    \includegraphics[width=\linewidth]{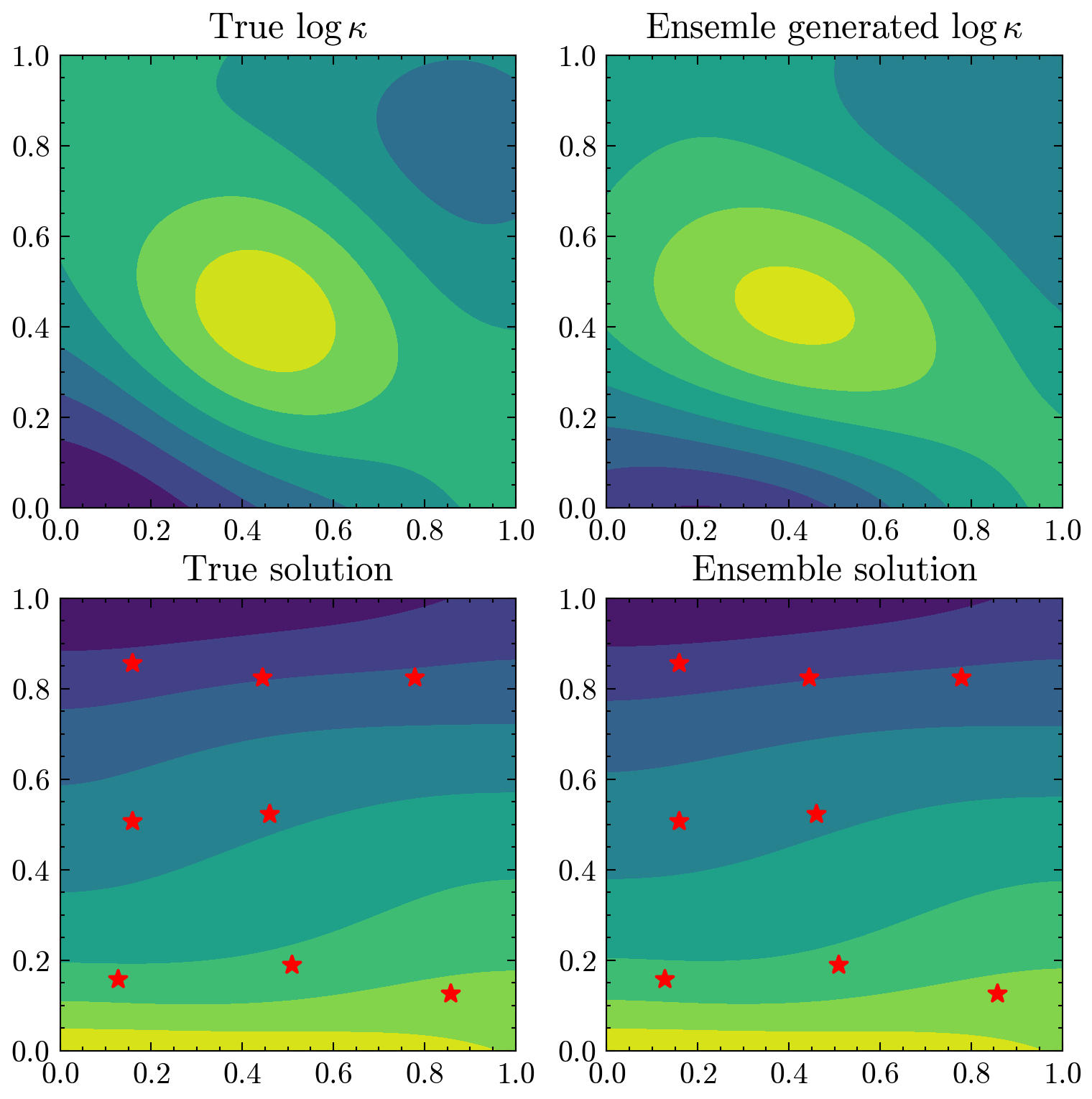}
    \caption{PDE coefficient and solution: true (left) and reconstructed using Flow Matching (right)}
    \label{fig:inv_perm}
\end{minipage}
\end{figure}

The true solution of the ODE system and the reconstructed parameter distribution, obtained using only four observation points, are illustrated in Figure \ref{fig:odes_sols}.

\begin{table}[!h]
    \centering
    \caption{The relative inference error of the trained model for two numerical experiments} \label{tab:seir_permeability_results}
    \begin{tabular}{rcc}
        \toprule
        \bfseries N & \bfseries SEIR Problem~\ref{subsec:seir}& \bfseries Permeability Field~\ref{subsec:inv_pde_3point} \\
        \midrule
        4 & $2.80\% \pm 1.37\%$ & $17.80\% \pm 1.99\%$ \\
        5 & $2.15\% \pm 0.99\%$ & $16.86\% \pm 1.76\%$ \\
        6 & $1.97\% \pm 0.91\%$ & $7.21\% \pm 1.26\%$ \\
        7 & $1.59\% \pm 0.75\%$ & $7.48\% \pm 1.23\%$ \\
        8 & $1.48\% \pm 0.71\%$ & $2.75\% \pm 0.60\%$ \\
        \bottomrule
    \end{tabular}
\end{table}


We compare our method against the Metropolis-Hastings MCMC (MH-MCMC) algorithm, running it with sufficient iterations to match the error levels shown in Table \ref{tab:seir_permeability_results}. The results for the SEIR problem are presented in Table \ref{tab:mcmc_for_problems}.

\begin{table}[ht]
\centering
\caption{Relative errors for numerical experiments using MCMC}
\label{tab:mcmc_for_problems}
\begin{tabular}{@{}ccccc@{}}
\toprule
\multicolumn{1}{l}{\multirow{2}{*}{\bfseries N }} & \multicolumn{2}{c}{\bfseries SEIR Problem} & \multicolumn{2}{c}{\bfseries Permeability field} \\ \cmidrule(l){2-5} 
\multicolumn{1}{l}{} & \bfseries N\(_{\text{sample}}\) & \bfseries Relative Error & \multicolumn{1}{c}{ \bfseries N\(_{\text{sample}}\) } & \multicolumn{1}{c}{\bfseries  Relative Error} \\ \midrule
2 & 15 000 & 31.39\% & 10 000  &  56.57\%\\
3 & 10 000 & 3.26\% & 10 000 &  39.40\%\\
4 & 5 000 & 3.24\% &  10 000 &  57.66\%\\
5 & 5 000 & 2.74\% & 10 000 & 55.71\% \\
6 & 5 000 & 1.64\% & 10 000 & 42.63\%  \\
7 & 5 000 & 4.07\% &  10 000 &  36.13\%\\
8 & 10 000 & 1.44\% &  5 000 &  60.02\% \\ \bottomrule
\end{tabular}
\end{table}

\paragraph{Comparison with MCMC} Comparing Tables \ref{tab:seir_permeability_results} and \ref{tab:mcmc_for_problems}, we observe that MCMC requires significantly more iterations to achieve similar accuracy as Conditional Flow Matching. This computational cost is particularly significant for elliptic PDEs. Even after 10,000 MCMC iterations, the error in solving the PDE remains above 30\% for 6 or more observations, while Conditional Flow Matching achieves 2-8\% error in these cases. The computational time difference is substantial: MCMC takes approximately 37 minutes for 10,000 samples, while Flow Matching requires only 1.08 seconds (on CPU) per inference using Algorithm \ref{alg:cfm_inference} for the permeability inversion problem. For the SEIR model, the CFM transformer takes approximately 0.22 seconds.

Our experiments highlight several key findings:

\begin{enumerate}
    \item Deep learning generative models can be effectively trained to handle variable-length inputs for solving Bayesian Inverse Problems. This is achieved through a Transformer architecture with Rotary Embeddings, enabling inference on sequences longer than those seen during training.
    \item Flow Matching successfully learns non-trivial paths from prior to target distributions, as evidenced by the nearly straight paths shown in Figure \ref{fig:flow_matching_pathes}.
    \item The ability to handle arbitrary numbers of observations proves highly beneficial, with Table \ref{tab:seir_permeability_results} showing consistent error reduction as the number of observations increases.
\end{enumerate}

Future research directions include combining Conditional Flow Matching with classical MCMC methods. Given MCMC's slow convergence, Flow Matching could serve as an improved prior distribution for MH-MCMC.

Additionally, Conditional Flow Matching shows promise for determining optimal experiment design parameters \(e\), which could further enhance its applicability to practical scientific applications.

\section{Limitations and Future Work}
\label{sec:limits}
The CFM approach is not the only generative 
modelling technique available. Alternative generative modeling techniques include normalizing flows, tensor methods\cite{koval2024tractableoptimalexperimentaldesign}, generative adversarial networks. We also need to establish the efficiency of the approach for high-dimensional Bayesian inverse problems, which will require modifications of the architecture and additional scaling of the datasets. CFM learns to sample from the distribution, whereas the Bayesian optimal experiment design requires the evaluation of the log-likelihood. Although it is in principle possible, we did not study the complexity and accuracy of the evaluation of the logarithm of the posterior distribution. 

Another line of work is to introduce the guidance to the CFM objective that is enforced by the forward model. Once we have obtained the estimate of the parameter $m$, we can check if it really fits the observations; a good question is how to modify the inference procedure to correct for possible errors. 

An important limitation is the study of the actual properties of the learned distribution. For the cases, 
when the model parameters are defined by the observations, the velocity field $v_t$ will likely not depend
on the noise, but only on $d, e$, effectively solving the regression problem of predicting $m_i$ from $d_i, e_i$. The usefulness and emergence of randomization with respect to the noise in CFM still needs to be studied.
There are two options for using it. First, we can sample different noises, get estimates of the model parameters and plot the distributions (see the Section on the SEIR model, where some of the parameters show greater variability). The second option is to pick samples that provide better fit to the data. This still needs to be studied, since in some of the experiments we found differences in the distribution provided by MCMC and CFM. The question has to be studied in more details.

\section{Code and Availability}
\label{sec:avail}

Technical training details (architectures, learning rates, etc.) are
given in Appendix \ref{app:tech}. The code is written using \texttt{PyTorch} framework and is publicly available at

\begin{center}
    \url{https://github.com/FlowMatchingInverseProblems/Bayesian-Inverse-Meet-FM}
\end{center}

\section{Conclusions}

We believe that our method is quite universal and can be adapted to a large number of problems in a short time when the problem is reduced to a standard Bayesian inverse problem formulation, since it can learn complex nonlinear distributions. Another advantage is the possibility of using an input that is not fixed in terms of the number of observations, where increasing the number of observed points improves accuracy in recovering the solution from the generated parameters. Finally, we can use the learned distribution to do Bayesian optimal experiment design.

\section*{References}
\bibliography{refs}


\newpage

\appendix

\section{Numerical Experiments Problem Statements}

\subsection{Simple nonlinear model}

In our experiments, we used the following forward model from \cite{koval2024tractableoptimalexperimentaldesign}:

\begin{equation*}
    d(e,m) = e^2 m^3 + m \exp \left(-\left| 0.2 - e \right|\right) + \eta
\end{equation*}

where $\eta$ follows a known noise distribution, specifically $\mathcal{N}(0, \sigma^2)$. In the simplest example from \cite{koval2024tractableoptimalexperimentaldesign}, the model parameter $m$ is one-dimensional, uniformly distributed on $[0, 1]$. The experiment parameter $e$ is also one-dimensional from $[0, 1]$ and uniformly distributed. We generate random triples $(d_i, m_i, e_i)$ by:

\begin{itemize}
    \item Sampling $m$ from $\mathcal{U}[0, 1]$
    \item Sampling $e$ from $\mathcal{U}[0, 1]$
    \item Sampling noise $\eta$ from the noise distribution
    \item Computing $d = f(m, e) + \eta$
\end{itemize}

After sampling, we obtain a dataset in the form of an $N \times k$ matrix, where $k = 3$. These are samples from the \textbf{joint distribution} $\pi(d, m, e)$. The prior distribution for training conditional flow matching was a simple uniform distribution \(m_0 \sim \mathcal{U}[0, 1]\).

\begin{figure}[htb]
\begin{center}
\centerline{\includegraphics[width=.5\columnwidth]{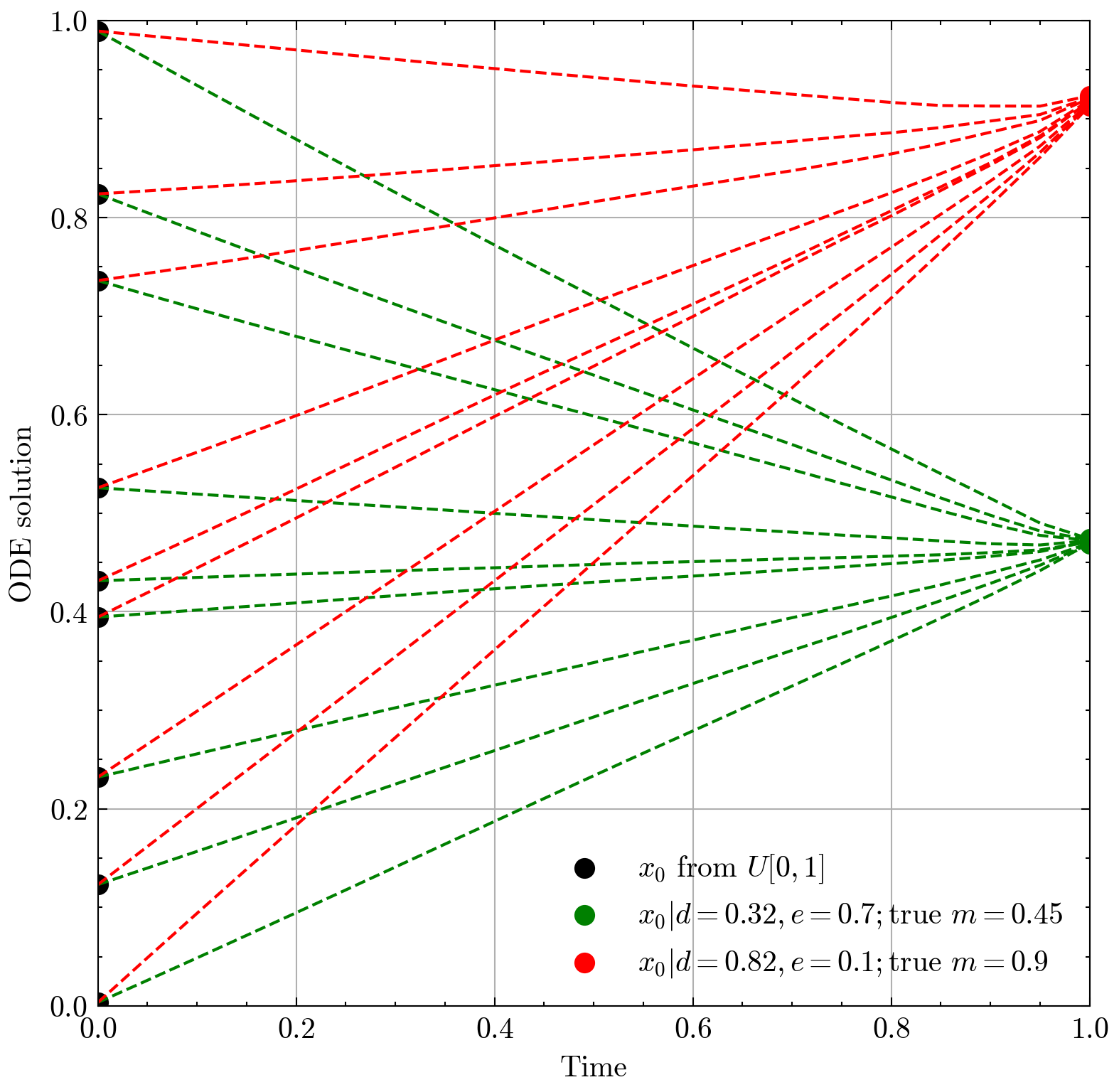}}
\caption{Generation paths of variable \(m\) conditional on different \(d\), \(e\) from prior uniform distribution}
\label{fig:flow_matching_pathes}
\end{center}
\end{figure}

\subsection{SEIR disease model}

Following \citep{koval2024tractableoptimalexperimentaldesign}, we adopt the SEIR model, which assumes a constant population size and is described by the following system of ordinary differential equations:

\begin{align*}
  \frac{dS}{dt} & = -\beta(t)SI, \frac{dE}{dt} = \beta(t) S I - \alpha E \\ 
  \frac{dI}{dt} &= \alpha E - \gamma(t)I, \frac{dR}{dt} = \gamma(t) I
\end{align*}

where $S(t)$, $E(t)$, $I(t)$, $R(t)$ denote the fractions of susceptible, exposed, infected, and removed individuals at time $t$, respectively. These are initialized with $S(0)=99$, $E(0)=1$, $I(0)=R(0) = 0$.

The parameters to be estimated are $\beta(t)$, $\alpha$, $\gamma^r$, and $\gamma^d(t)$, where the constants $\alpha$ and $\gamma^r$ denote the rates of susceptibility to exposure and infection to recovery, respectively. To simulate the effect of policy changes or other time-dependent factors (e.g., quarantine and hospital capacity), the rates at which exposed individuals become infected and infected individuals perish are assumed to be time-dependent and parametrized as:

\begin{align*}
  \beta(t) &= \beta_1 + \frac{\tanh(7(t-\tau))}{2}(\beta_2 - \beta_1) \\ 
  \gamma(t) &= \gamma^r + \gamma^d(t)\\
  \gamma^d(t) &= \gamma^d_1 + \frac{\tanh(7(t-\tau))}{2}(\gamma^d_2 - \gamma^d_1)
\end{align*}

where the rates transition smoothly from initial rates ($\beta_1$ and $\gamma^d_1$) to final rates ($\beta_2$ and $\gamma^d_2$) around time $\tau > 0$.

We fix $\tau = 2.1$ over a time interval of $[0, 4]$. The experiment consists of choosing four time points $e = [a_1, a_2, a_3, a_4] \sim \mathcal{U}[1,3]$ to measure the number of infected and deceased individuals $d_i = [I_{e_i}, R_{e_i}]$ for $i \in [1,4]$ ($d \in \mathbf{R}^{2 \times 4}$). The goal is to optimally infer the 6 rates $\mathbf{m} = [\beta_1, \alpha, \gamma^r, \gamma^d_1, \beta_2, \gamma^d_2]$. After training an MLP and solving the flow matching problem, we learn a smooth transition from the distribution $\mathcal{U}[0,1]^6$ to the distribution $\hat{\mathbf{m}} \sim \rho(\mathbf{m} \vert \mathbf{e} , \mathbf{d})$.

To summarize the inputs and outputs:
\begin{itemize}
    \item $e = [a_1, a_2, a_3, a_4] \sim \mathcal{U}[1,3]$: random measurement times
    \item $d_i = [I_{e_i}, R_{e_i}]$ for $i \in [1,4]$ ($d \in \mathbf{R}^{2 \times 4}$): numbers of infected and deceased individuals
    \item $m = [\beta_1, \alpha, \gamma^r, \gamma^d_1, \beta_2, \gamma^d_2]$: ODE model parameters
\end{itemize}

Using $\hat{m}$, we can obtain the predicted dynamics of infected and deceased individuals $\hat{d}$. We measure accuracy using:

\begin{equation*}
    \varepsilon = \frac{\Vert d - \hat{d}\Vert_2}{\Vert d \Vert_2}
\end{equation*}

\subsection{Permeability field inversion}

\begin{equation*}
    - \nabla \cdot (\kappa \nabla u) = 0
\end{equation*}

with boundary conditions:

\begin{align*}
    u(x = 0, y) &= f(y, e_1) = \exp\left(-\frac{1}{2\sigma_w} (y - e_1)^2\right) \\ 
    u(x = 1, y) &= g(y, e_2) = - \exp\left(-\frac{1}{2\sigma_w} (y - e_2)^2\right)
\end{align*}

The equation is solved using the finite element (FE) method with second-order Lagrange elements on a mesh of size $h = \frac{1}{64}$ in each coordinate direction, where $\kappa$ is a custom 2D matrix. The discretization follows \cite{DOLGOV20122980}.

In this example, the inverse problem consists of estimating the spatially-dependent diffusivity field $\kappa$ given pressure measurements $u$ at pre-determined locations $(x_i, y_i) \in \Omega$. To ensure $\kappa$ is nonnegative, we impose a Gaussian prior on the log diffusivity, $m = \log(\kappa) \sim N(0, C_{pr})$, with covariance operator $C_{pr}$ defined using a squared-exponential kernel:

\begin{equation*}
    c(x, z) = \sigma_v^2 \exp \left[ \frac{-\|x - z\|^2}{2 \ell^2} \right] \quad \text{for } x, z \in \Omega,
\end{equation*}

with $\sigma_v = 1$ and $\ell^2 = 0.1$. Using a truncated Karhunen-Loève expansion of the unknown diffusivity field yields the approximation:

\begin{equation*}
   m(x, \mathbf{m}) \approx \sum_{i=1}^{n_m} m_i \sqrt{\lambda_i} \phi_i(x), 
\end{equation*}

where $\lambda_i$ and $\phi_i(x)$ denote the $i$-th largest eigenvalue and eigenfunction of $C_{pr}$, respectively, and the unknown coefficients $m_i \sim \mathcal{N}(0, 1)$. The Karhunen-Loève expansion is truncated after $n_m = 16$ modes, capturing 99 percent of the weight of $C_{pr}$.

The transformer architecture accommodates various input formats for this inverse problem. Here, in addition to the observed solution values, we use the coordinates of measurement points. The specific architecture is detailed in Figure \ref{fig:arch_scheme_2} in Appendix \ref{app:tech}.

The input consists of a vector of values $d$ of arbitrary length and two corresponding vectors of coordinates $x, y$. The final input is a matrix $\mathbf{D} = (\mathbf{d}, \mathbf{x}, \mathbf{y})^T$ with shape $(n,3)$.

\section{Technical details}
\label{app:tech}

The transformer architecture for two numerical experiments

\begin{figure}[ht]
  \centering
  \includegraphics[width=\linewidth]{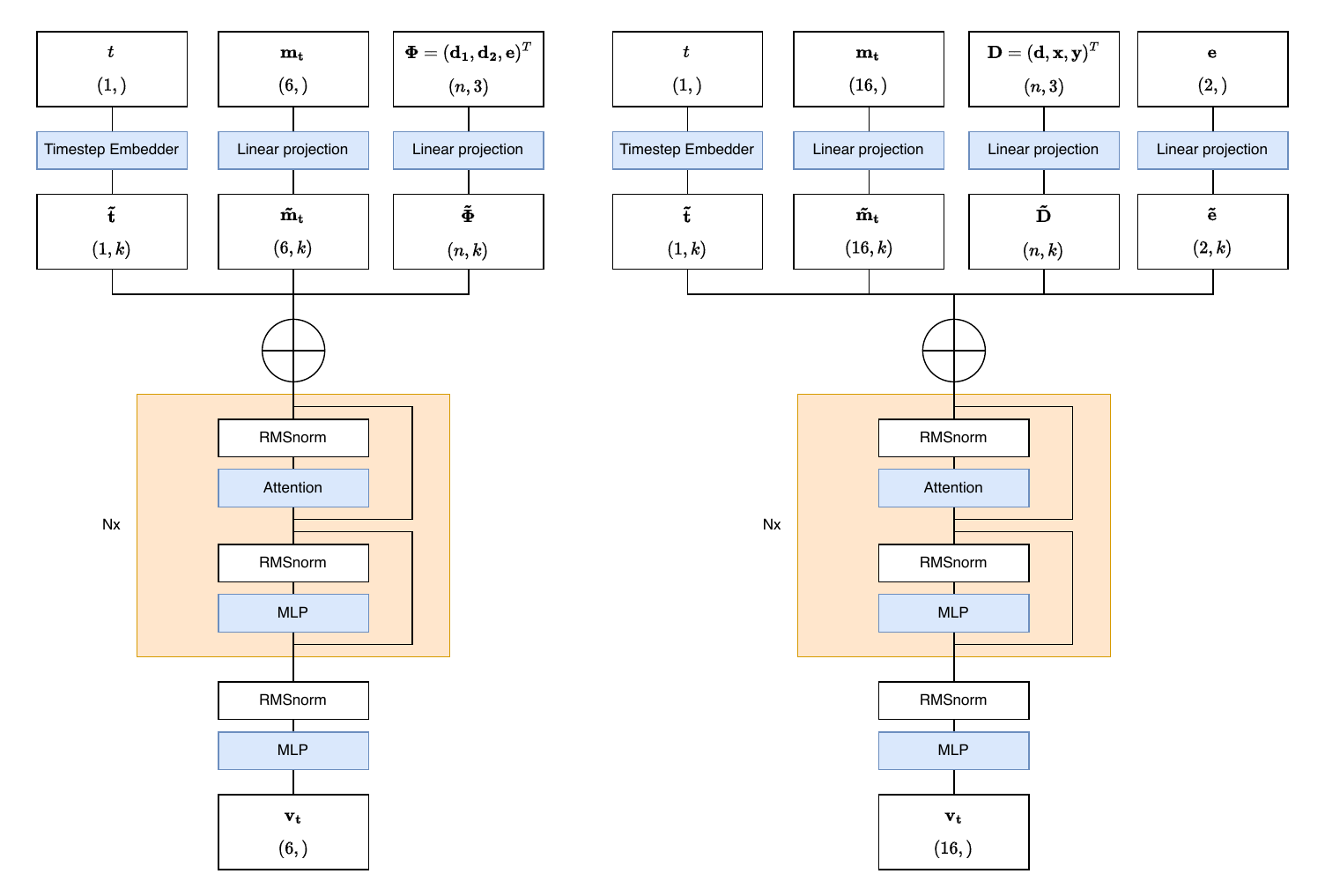}
\caption{Transformer architecture for \ref{subsec:seir} (left) and \ref{subsec:inv_pde_3point} (right)}
\label{fig:arch_scheme_2}
\end{figure}

\begin{table}[ht]
\centering
\caption{Hyperparameters for SEIR and Permeability Inversion tasks}
\begin{tabular}{lcc}
\toprule
\textbf{Parameter} & \textbf{SEIR} & \textbf{Permeability Inversion} \\
\midrule
learning\_rate & 8e-4 & 3e-4 \\
n\_emb         & 32   & 32   \\
n\_head        & 4    & 4    \\
n\_layer       & 6    & 4    \\
\bottomrule
\end{tabular}
\label{tab:hyperparams}
\end{table}

\section{Algorithms}

\subsection{Conditional Flow Matching}

This section provides pseudocode for the core training and inference procedures used in our Conditional Flow Matching (CFM) framework. These algorithms form the backbone of our method for solving inverse problems in various scientific settings.

Algorithm~\ref{alg:cfm} details the training procedure for the conditional flow model. Given a dataset of paired samples and conditioning information, the model is trained to approximate the velocity field that defines an interpolation between prior and posterior samples. The training objective minimizes the squared error between the predicted velocity and the ground-truth velocity vector defined by the linear interpolation between samples.

\begin{algorithm}[htb]
\DontPrintSemicolon
\SetAlgoLined
\KwIn{
    Dataset of paired samples \((m_1, e, d)\), neural network model \(\mathbf{v}_\theta(t, m, e, d)\), conditioning data \(e\) and \(d\), time \(t \sim \text{Uniform}(0, 1)\), number of epochs \(N_{\text{epoch}}\)
}
\KwOut{Trained conditional flow model \(\mathbf{v}_\theta(t, m, e, d)\)}

\For{\(1\) \textbf{to} \(N_{\text{epoch}}\)}{

\For{each minibatch of samples \((m_0, m_1)\)}{
    \(t \sim \mathcal{U}(0, 1)\) \tcp*[r]{Sample t}

    \(m_0 \sim \text{prior distribution}\)

    \(m_t \gets t \cdot m_1 + (1 - t) \cdot m_0\)

    Compute the target velocity: \(u_t \gets m_1 - m_0\)

    Predict the velocity: \(v_t \gets \mathbf{v}(t, m_t, e, d)\)
    
    Compute the loss: \(\mathcal{L}(\theta) \gets \mathbb{E} \left[ \Vert v_t - u_t \Vert^2 \right]\)

    Compute gradients: $\nabla_\theta \mathcal{L}(\theta)$\;

    Update \(\theta\) using the optimizer and \(\nabla_\theta \mathcal{L}(\theta)\)\;
}
}
\Return{\(\mathbf{v}_\theta(t, x, e, d)\)}

\caption{Conditional Flow Matching Training Algorithm}
\label{alg:cfm}
\end{algorithm}

Algorithm~\ref{alg:cfm_inference} presents the inference procedure. After training, the model is used to define a deterministic flow by solving an ordinary differential equation (ODE) starting from a sample from the prior distribution. The terminal state of this ODE corresponds to a sample from the conditional distribution given the observations and experimental conditions.

Together, these two procedures enable the model to learn and sample from complex conditional distributions without relying on stochastic sampling or iterative optimization during inference.

\begin{algorithm}[H]
\DontPrintSemicolon
\SetAlgoLined
\KwIn{
    Trained CFM model \(\mathbf{v}_\theta(t, x)\), conditioning data \(e\) and \(d\), initial sample \(x_0\), experiment parameters \(e\), arbitrary observations \(d\)
}
\KwOut{Generated parameters \(m\)}

\(x(t = 0) \sim \text{prior distribution}\)

\(x(t = 1)  \gets \text{Solution}\frac{d x}{d t} = \mathbf{v}_\theta(t, x_t, e, d)\)\;

\Return{\(x(t = 1)\)}

\caption{Conditional Flow Matching Inference Algorithm}
\label{alg:cfm_inference}
\end{algorithm}

\end{document}